\documentclass[runningheads]{llncs}
\usepackage{graphicx}

\usepackage{tikz}
\usepackage{comment}
\usepackage{amsmath,amssymb} 
\usepackage{color}
\usepackage{booktabs}
\usepackage{bm}
\usepackage{multirow}
\usepackage{wrapfig}

\usepackage[pagebackref,breaklinks,colorlinks]{hyperref}
\usepackage[capitalize]{cleveref}

\usepackage{pifont}
\newcommand{\cmark}{\ding{51}}%
\newcommand{\xmark}{\ding{55}}%

\usepackage{amsmath,amsfonts,bm}

\def\eqref#1{equation~\ref{#1}}

\def\1{\bm{1}}

\DeclareMathAlphabet{\mathsfit}{\encodingdefault}{\sfdefault}{m}{sl}
\SetMathAlphabet{\mathsfit}{bold}{\encodingdefault}{\sfdefault}{bx}{n}

\newcommand{\R}{\mathbb{R}}

\begin{document}
\pagestyle{headings}
\mainmatter
\def\ECCVSubNumber{100}  

\title{AdaFocusV3: On Unified Spatial-temporal Dynamic Video Recognition} 

\titlerunning{AdaFocusV3} 
\authorrunning{Y. Wang \emph{et al.}} 
\author{Yulin Wang\inst{1}\thanks{Equal contributions.\ \ \ \ \ \ \ \ $^\dagger$Corresponding Authors.} \and
Yang Yue\inst{1}\inst{\star} \and
Xinhong Xu\inst{1} \and
Ali Hassani\inst{2} \and
Victor Kulikov\inst{3} \and
Nikita Orlov\inst{3} \and
Shiji Song\inst{1} \and
Humphrey Shi\inst{2, 3}{$^\dagger$} \and
Gao Huang\inst{1, 4}{$^\dagger$}
}

\institute{
    Department of Automation, BNRist, Tsinghua University \and
    University of Oregon \and
    Picsart AI Research (PAIR) \and
    Beijing Academy of Artificial Intelligence (BAAI) \\
    \email{\{wang-yl19, yueyang22\}@mails.tsinghua.edu.cn, shihonghui3@gmail.com, gaohuang@tsinghua.edu.cn} 
}

\maketitle

\begin{abstract}

Recent research has revealed that reducing the \emph{temporal} and \emph{spatial} redundancy are both effective approaches towards efficient video recognition, \emph{e.g.}, allocating the majority of computation to a task-relevant subset of frames or the most valuable image regions of each frame. However, in most existing works, either type of redundancy is typically modeled with another absent. This paper explores the unified formulation of spatial-temporal dynamic computation on top of the recently proposed AdaFocusV2 algorithm, contributing to an improved AdaFocusV3 framework. Our method reduces the computational cost by activating the expensive high-capacity network only on some small but informative 3D video cubes. These cubes are cropped from the space formed by frame height, width, and video duration, while their locations are adaptively determined with a light-weighted policy network on a per-sample basis. At test time, the number of the cubes corresponding to each video is dynamically configured, \emph{i.e.}, video cubes are processed sequentially until a sufficiently reliable prediction is produced. Notably, AdaFocusV3 can be effectively trained by approximating the non-differentiable cropping operation with the interpolation of deep features. Extensive empirical results on six benchmark datasets (\emph{i.e.}, ActivityNet, FCVID, Mini-Kinetics, Something-Something V1\&V2 and Diving48) demonstrate that our model is considerably more efficient than competitive baselines.










\end{abstract}

\section{Introduction}

Modern deep networks have reached or even surpassed human-level performance on large-scale video recognition benchmarks \cite{tran2015learning,feichtenhofer2016convolutional,carreira2017quo,hara2018can,feichtenhofer2019slowfast,feichtenhofer2020x3d,arnab2021vivit}. Such a remarkable success is in general fueled by the rapid development of large models with high computational demands at inference time. However, the practical application of these resource-hungry models may be challenging, despite their state-of-the-art accuracy. For example, in real-world scenarios like YouTube video recommendation \cite{davidson2010youtube,deldjoo2016content,gao2017unified}, video-based surveillance \cite{collins2000system,chen2019distributed} and content-based searching engines \cite{ikizler2007searching}, deploying computationally intensive networks significantly increases power consumption, system latency or carbon emission, all of which should be minimized due to economic and environmental concerns.

To address this issue, a number of recent works focus on reducing the inherent \emph{temporal} or \emph{spatial} redundancy in video analysis. For instance, for the former, several algorithms have been proposed to dynamically identify the most informative frames and allocate the majority of computation to them, yielding improvements in the overall efficiency \cite{yeung2016end,wu2019adaframe,gao2020listen,korbar2019scsampler,meng2020ar,sun2021dynamic,lin2022ocsampler}. For the latter, the recently proposed adaptive focus networks (AdaFocus) \cite{Wang_2021_ICCV,wang2021adafocus} develop dynamic models to adaptively attend to the task-relevant regions of each video frame, which considerably reduces the computational cost without sacrificing accuracy. Although both directions have been proven feasible, how to model \emph{spatial} and \emph{temporal} redundancy jointly and realize highly efficient spatial-temporal dynamic computation is still an under-explored topic. The preliminary results presented by AdaFocusV2+ \cite{wang2021adafocus} have revealed their favorable compatibility. However, AdaFocusV2+ only considers them separately in independent modules.


In this paper, we seek to explore the unified formulation that by design aims to reduce the spatial-temporal redundancy simultaneously, and thus study whether such general frameworks can lead to a more promising approach towards efficient video recognition. To implement this idea, we propose an AdaFocusV3 network. In specific, given an input video, our method first takes a quick and cheap glance at it with a light-weighted global network. A policy network will be learned on top of the obtained global information, whose training objective is to capture the most task-relevant parts of the video directly in the 3D space formed by the frame height/width and the time. This is achieved by localizing a sequence of 3D cubes inside the original video. These smaller but informative video cubes will be progressively processed using a high-capacity but computationally more expensive local network for learning discriminative representations. This procedure is dramatically more efficient than processing the whole video due to the reduced size of the cubes. As a consequence, the computation is unevenly allocated across both spatial and temporal dimensions, resulting in considerable improvements in the overall efficiency with the maximally preserved accuracy. 


It is noteworthy that the operation of selecting 3D cubes from the videos is not inherently differentiable. To enable the effective training of our AdaFocusV3 framework, we propose a gradient estimation algorithm for the policy network based on deep features. Our solution significantly outperforms the pixel-based counterpart proposed by AdaFocusV2 in our problem.

In addition, an intriguing property of AdaFocusV3 is that it naturally facilitates the adaptive inference. To be specific, since the informative video cubes are processed in a sequence, the inference procedure can be dynamically terminated once the model is able to produce sufficiently reliable predictions, such that further redundant computation on relatively ``easier'' samples can be saved. This paradigm also allows AdaFocusV3 to adjust the computational cost online without additional training (by simply adapting the early-termination criterion). Such flexibility of our method enables it to make full use of the fluctuating computational resources or minimize the power consumption with varying performance targets. Both of them are the practical requirements of many real-world applications (\emph{e.g.}, searching engines and mobile apps).

We validate the effectiveness of AdaFocusV3 on widely-used video recognition benchmarks. Empirical results show that AdaFocusV3 achieves a new state-of-the-art performance in terms of computational efficiency. For example, when obtaining the same accuracy, AdaFocusV3 has up to {2.6}$\times$ less Multiply-Add operations compared to the recently proposed OCSampler \cite{lin2022ocsampler} algorithm.

\section{Related Works}

\textbf{Video recognition.}
Recent years have witnessed a significant improvement of the test accuracy on large-scale automatic video recognition benchmarks \cite{caba2015activitynet,TPAMI-fcvid,kay2017kinetics,goyal2017something,materzynska2019jester}. This remarkable progress is largely ascribed to the rapid development of video representation learning backbones \cite{tran2015learning,wang2016temporal,carreira2017quo,hara2018can,feichtenhofer2019slowfast,lin2019tsm,feichtenhofer2020x3d,arnab2021vivit}. The majority of these works focus on modeling the temporal relationships among different frames, which is a key challenge in video understanding. Towards this direction, a representative approach is leveraging the spatial-temporal information simultaneously by expanding the 2-D convolution to 3D space \cite{tran2015learning,carreira2017quo,hara2018can,feichtenhofer2020x3d}. Another line of works propose to design specialized temporal-aware architectures on top of 2-D deep networks, such as adding recurrent networks \cite{donahue2015long,li2018recurrent,yue2015beyond}, temporal feature averaging \cite{wang2016temporal}, and temporal channel shift \cite{lin2019tsm,sudhakaran2020gate,fan2020rubiksnet,meng2021adafuse}. Some other works model short-term and long-term temporal relationships separately using two-stream networks \cite{feichtenhofer2016convolutional,feichtenhofer2017spatiotemporal,feichtenhofer2019slowfast,gong2021searching}. Besides, since processing videos with deep networks is computationally intensive, a variety of recent works have started to focus on developing efficient video recognition models \cite{tran2018closer,zolfaghari2018eco,tran2019video,luo2019grouped,liu2020teinet,liu2021tam}. 

\textbf{Temporal dynamic networks.}
A straightforward approach for facilitating efficient video representation learning is to leverage the inherent temporal redundancy within videos. To be specific, not all video frames are equivalently relevant to a given task, such that ideally a model should dynamically identify less informative frames and allocate less computation to them correspondingly \cite{han2021dynamic}. In the context of video recognition, a number of effective approaches have been proposed under this principle \cite{yeung2016end,wu2019adaframe,gao2020listen,korbar2019scsampler,wu2019multi,meng2020ar,ghodrati2021frameexit,kim2021efficient,sun2021dynamic,lin2022ocsampler}. For example, OCSampler \cite{lin2022ocsampler} proposes a one-step framework, learning to select task-relevant frames with reinforcement learning. VideoIQ \cite{sun2021dynamic}  processes the frames using varying precision according to their importance. FrameExit \cite{ghodrati2021frameexit} performs early-termination at test time after processing sufficiently informative frames. AdaFocusV3 is more general and flexible than these works since we simultaneously model both spatial and temporal redundancy.

\textbf{Spatial dynamic networks.}
In addition to the temporal dimension, considerable spatial redundancy exists when processing video frames. As a matter of fact, many works have revealed that deep networks can effectively extract representations from the image-based data via attending to a few task-relevant image regions \cite{figurnov2017spatially,NeurIPS2020_7866,xie2020spatially,verelst2020dynamic}. Recently, the effectiveness of this paradigm has been validated in video recognition as well. As representative works, AdaFocus network V1\&V2 \cite{Wang_2021_ICCV,wang2021adafocus} propose to infer the expensive high-capacity network only on a relatively small but informative patch of each frame, which is adaptively localized with a policy network, yielding a favorable accuracy-speed trade-off. An important advantage of the spatial-based approaches is that they are orthogonal to the temporal-adaptive methods. Nonetheless, in existing works, the spatial and temporal redundancy is usually modeled using independent algorithms with separate network architectures, and combined straightforwardly (\emph{e.g.}, AdaFocusV1\&V2 \cite{Wang_2021_ICCV,wang2021adafocus}). This paper assumes that such a naive implementation may lead to sub-optimal formulation, and proposes a unified AdaFocusV3 framework to consider spatial-temporal redundancy simultaneously, which achieves significantly higher computational efficiency.

\section{Method}
\label{sec:method}

As aforementioned, existing efficient video recognition approaches usually consider the temporal or spatial redundancy of videos with another absent. We assume that such isolated modeling of the two types of redundancy may lead to sub-optimal formulation, and propose a unified AdaFocusV3 framework. AdaFocusV3 learns to perform spatial-temporal dynamic computation simultaneously, yielding significantly improved computational efficiency compared with the state-of-the-art baselines. In the following, we introduce the details of our method.

\begin{figure}[t]
    \begin{center}
    \centerline{\includegraphics[width=0.925\columnwidth]{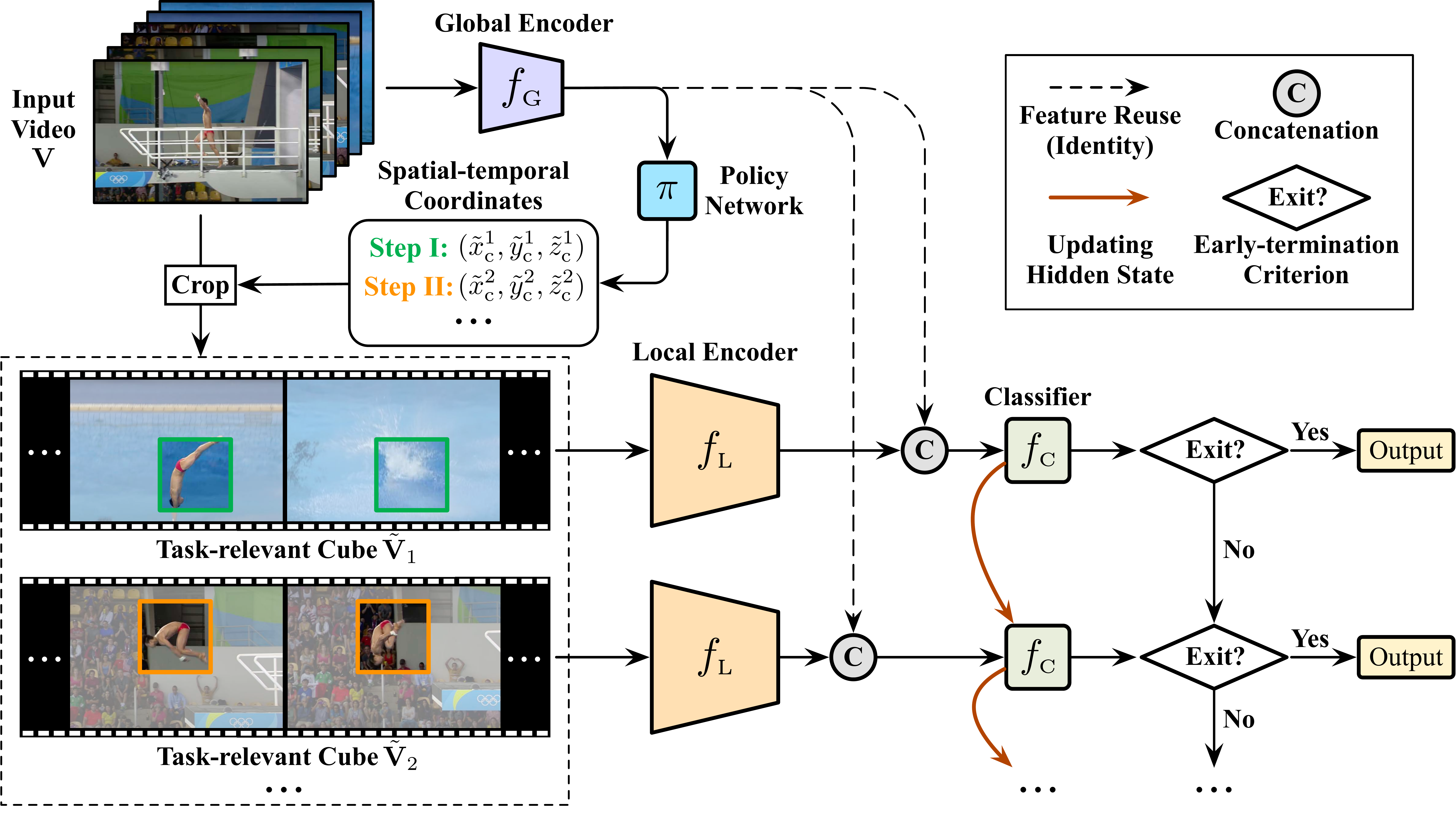}}
    \caption{\textbf{Inference procedure of AdaFocusV3.} 
        The global encoder $f_{\textnormal{G}}$ first takes a quick glance at each input video in the holistic view. A policy network $\bm{\pi}$ is learned on top of $f_{\textnormal{G}}$ to capture the most task-relevant parts of the video with 3D spatial-temporal cubes. These cubes are processed by the high-capacity and accurate local encoder $f_{\textnormal{L}}$ for extracting discriminative deep features. A classifier $\bm{\pi}$ aggregates all the obtained information and dynamically outputs the prediction. Notably, the inference will be terminated once a convincing prediction (\emph{e.g.}, with sufficiently low entropy or being adequately confident) has been produced.
    \label{fig:overview}
    }
    \end{center}
\end{figure}

\subsection{Overview}
\label{sec:overview}

We first describe the inference and training pipelines of AdaFocusV3, laying the basis for the detailed introduction to the network architecture, the training algorithm and the implementation details.


\textbf{Inference.}
We start by introducing the inference procedure of AdaFocusV3, which is shown in Figure \ref{fig:overview}. Given an input video $\mathbf{V}\!\in\!\R^{H\times W\times T}$ with $T$ frames of $H\!\times\!W$ images (here we omit the RGB channels for simplicity), we first process it with a light-weighted global encoder $f_{\textnormal{G}}$. The aim of this step is to obtain the coarse global information cheaply with low computational cost. Then the output features of $f_{\textnormal{G}}$ are fed into a policy network $\bm{\pi}$, which is trained to capture the most task-relevant parts of $\mathbf{V}$ for extracting finer representations. In specific, the outputs of $\bm{\pi}$ localize a sequence of 3D cubes $\{\tilde{\mathbf{V}}_1, \tilde{\mathbf{V}}_2, \dots\}$ with the size $H'\!\times\!W'\!\times\!T'$ ($H'\!<\!H, W'\!<\!W, T'\!<\!T$), and these cubes will be processed by a local encoder $f_{\textnormal{L}}$. Note that $f_{\textnormal{L}}$ is designed to be high-capacity, accurate and computationally more expensive. Considerable redundant computation can be saved by only activating $f_{\textnormal{L}}$ on top of the selected informative inputs with reduced size rather than the whole video.

At last, a classifier $f_{\textnormal{C}}$ aggregates the features of all the previous inputs to produce a prediction. Importantly, the contribution of $\{\tilde{\mathbf{V}}_1, \tilde{\mathbf{V}}_2, \dots\}$ to recognition are formulated to be descending in the sequence (via training, details described later). They are progressively processed by $f_{\textnormal{L}}$ and $f_{\textnormal{C}}$, while a softmax prediction $\bm{p}_t$ will be retrieved from $f_{\textnormal{C}}$ after seeing every video cube $\tilde{\mathbf{V}}_t$. In other words, ideally, AdaFocusV3 always allocates the computation first to the most important video contents in terms of the task. The major insight behind this design is to facilitate the adaptive inference, \emph{i.e.}, the inference can be terminated once $\bm{p}_t$ is sufficiently reliable and thus avoids further redundant computation. In the implementation, we adopt an entropy-based early-termination criterion (see Section \ref{sec:implementation_details} for details). 

\textbf{Training.}
AdaFocusV3 is trained by minimizing the sum of the loss corresponding to all the predictions from $f_{\textnormal{C}}$, namely
\begin{equation}
    \label{eq:training_objective}
        \mathop{\textnormal{minimize}}_{f_{\textnormal{G}}, f_{\textnormal{L}}, f_{\textnormal{C}}, \pi}\ \ \mathcal{L}=\mathop{\mathbb{E}}_{(\mathbf{V}, y) \in \mathcal{D}_{\textnormal{train}}}
    \left[
        \sum\nolimits_{t} L_{\textnormal{CE}}(\bm{p}_t, y)
    \right], 
\end{equation}
where $y$ is the label of $\mathbf{V}$, $\mathcal{D}_{\textnormal{train}}$ is the training set and $L_{\textnormal{CE}}(\cdot)$ denotes the standard cross-entropy loss function. Intuitively, solving problem (\ref{eq:training_objective}) will learn a $\bm{\pi}$ that enables the model to produce correct predictions with as fewer inputs as possible. The selection policy is trained to usually find the most beneficial cubes for the current model at each step.

\subsection{Network Architecture}

\textbf{Global encoder $f_{\textnormal{G}}$ and local encoder $f_{\textnormal{L}}$}
are both deep networks (\emph{e.g.}, ConvNets and vision Transformers) that map the input video frames to the deep feature space. However, $f_{\textnormal{G}}$ is exploited to take a quick glance at the videos, such that the policy network $\bm{\pi}$ can be activated to localize the informative video cubes. As a consequence, light-weighted architectures should be adopted. On the other hand, $f_{\textnormal{L}}$ is designed to extract accurate and discriminative representations from the selected important inputs. Therefore, it allows deploying computationally intensive and high-capacity models.

\textbf{Policy network $\bm{\pi}$}
receives the global feature maps produced by $f_{\textnormal{G}}$, and determines the locations of the video cubes to attend to. Notably, here we assume all the frames of $\mathbf{V}$ are fed into $f_{\textnormal{G}}$ at the same time, since AdaFocusV3 seeks to reduce both the temporal and spatial redundancy simultaneously. In contrast, AdaFocusV2/V1 \cite{Wang_2021_ICCV,wang2021adafocus} by design processes the videos frame by frame. To this end, we do not follow their recurrent design of $\bm{\pi}$. On the contrary, we directly process the global features of the whole video, and obtain the centre coordinates of $\{\tilde{\mathbf{V}}_1, \tilde{\mathbf{V}}_2, \dots\}$ at one time.


\textbf{Classifier $f_{\textnormal{C}}$}
aggregates the information from $\{\tilde{\mathbf{V}}_1, \tilde{\mathbf{V}}_2, \dots\}$, and produces the final prediction correspondingly. The architecture of $f_{\textnormal{C}}$ may have various candidates, including recurrent networks \cite{hochreiter1997long,cho-etal-2014-learning,Wang_2021_ICCV}, frame-wise prediction averaging \cite{lin2019tsm,meng2020ar,meng2021adafuse}, and accumulated feature pooling \cite{ghodrati2021frameexit,wang2021adafocus}. Besides, we adopt the efficient feature reuse mechanism proposed by \cite{Wang_2021_ICCV,wang2021adafocus}, where the outputs of both $f_{\textnormal{G}}$ and $f_{\textnormal{L}}$ are leveraged by $f_{\textnormal{C}}$ (depicted using dashed lines in Figure \ref{fig:overview}).

\subsection{Training Algorithm}

\textbf{Naive implementation.}
An obstacle towards solving problem (\ref{eq:training_objective}) lies in that, cropping the cubes $\{\tilde{\mathbf{V}}_1, \tilde{\mathbf{V}}_2, \dots\}$ from the video $\mathbf{V}$ is not an inherently differentiable operation. Consequently, it is nontrivial to apply the standard back-propagation-based training algorithm. To address this issue, a straightforward approach is to leverage the interpolation-based cropping mechanism proposed by AdaFocusV2 \cite{wang2021adafocus}. Formally, given the centre coordinates $(\tilde{x}^t_{\textnormal{c}}, \tilde{y}^t_{\textnormal{c}}, \tilde{z}^t_{\textnormal{c}})$ of $\tilde{\mathbf{V}}_t$, we can obtain $\tilde{\mathbf{V}}_t$ through trilinear interpolation. In this way, the value of any pixel $\tilde{v}^t_{i,j,k}$ in $\tilde{\mathbf{V}}_t$ is calculated as the weighted combination of its surrounded eight adjacent pixels in $\mathbf{V}$, \emph{i.e.},
\begin{equation}
    \label{eq:trilinear}
    \tilde{v}^t_{i,j,k} = \textnormal{Trilinear}\left(\mathbf{V}, (\tilde{x}^t_{i,j,k}, \tilde{y}^t_{i,j,k}, \tilde{z}^t_{i,j,k})\right), 
\end{equation}
where $(\tilde{x}^t_{i,j,k}, \tilde{y}^t_{i,j,k}, \tilde{z}^t_{i,j,k})$ corresponds to the horizontal, vertical and temporal coordinates of $\tilde{v}^t_{i,j,k}$ in $\mathbf{V}$.

In back-propagation, it is easy to obtain the gradients of the loss $\mathcal{L}$ with respect to $\tilde{v}^t_{i,j,k}$, \emph{i.e.}, ${\partial\mathcal{L}}/{\partial\tilde{v}^t_{i,j,k}}$. With Eq. (\ref{eq:trilinear}), we further have
\begin{equation}
    \label{eq:bp}
        \frac{\partial\mathcal{L}}{\partial\tilde{x}^t_{\textnormal{c}}} \!=\!\! \sum_{i,j,k}\! \frac{\partial\mathcal{L}}{\partial\tilde{v}^t_{i,j,k}}
        \frac{\partial\tilde{v}^t_{i,j,k}}{\partial\tilde{x}^t_{i,j,k}}, \ 
        \frac{\partial\mathcal{L}}{\partial\tilde{y}^t_{\textnormal{c}}} \!=\!\! \sum_{i,j,k}\! \frac{\partial\mathcal{L}}{\partial\tilde{v}^t_{i,j,k}}
        \frac{\partial\tilde{v}^t_{i,j,k}}{\partial\tilde{y}^t_{i,j,k}}, \ 
        \frac{\partial\mathcal{L}}{\partial\tilde{z}^t_{\textnormal{c}}} \!=\!\! \sum_{i,j,k}\! \frac{\partial\mathcal{L}}{\partial\tilde{v}^t_{i,j,k}}
        \frac{\partial\tilde{v}^t_{i,j,k}}{\partial\tilde{z}^t_{i,j,k}}.
\end{equation}
Note that Eq. (\ref{eq:bp}) leverages the fact that the geometric relationship between $(\tilde{x}^t_{i,j,k}, \tilde{y}^t_{i,j,k}, \tilde{z}^t_{i,j,k})$ and $(\tilde{x}^t_{\textnormal{c}}, \tilde{y}^t_{\textnormal{c}}, \tilde{z}^t_{\textnormal{c}})$ is always fixed once the size of $\tilde{\mathbf{V}}_t$ is fixed, namely we have
\begin{equation}
    \label{eq:bp_2}
    \frac{\partial\tilde{v}^t_{i,j,k}}{\partial\tilde{x}^t_{\textnormal{c}}}\!=\!\frac{\partial\tilde{v}^t_{i,j,k}}{\partial\tilde{x}^t_{i,j,k}},\ 
    \frac{\partial\tilde{v}^t_{i,j,k}}{\partial\tilde{y}^t_{\textnormal{c}}}\!=\!\frac{\partial\tilde{v}^t_{i,j,k}}{\partial\tilde{y}^t_{i,j,k}},\ 
    \frac{\partial\tilde{v}^t_{i,j,k}}{\partial\tilde{z}^t_{\textnormal{c}}}\!=\!\frac{\partial\tilde{v}^t_{i,j,k}}{\partial\tilde{z}^t_{i,j,k}}.
\end{equation}
Given that $\tilde{x}^t_{\textnormal{c}}$, $\tilde{y}^t_{\textnormal{c}}$ and $\tilde{z}^t_{\textnormal{c}}$ are the outputs of $\bm{\pi}$, the regular back-propagation are capable of proceeding through Eq. (\ref{eq:bp}) to compute the gradients of all the parameters and update the model.

In essence, calculating the gradients with Eqs. (\ref{eq:trilinear}-\ref{eq:bp_2}) follows the following logic. We change the location of $\tilde{\mathbf{V}}_t$ in an infinitesimal quantity, and measure the consequent changes on the value of each pixel in $\tilde{\mathbf{V}}_t$ via Eq. (\ref{eq:trilinear}). These changes affect the inputs of $f_{\textnormal{L}}$, and accordingly, have an impact on the final optimization objective. 

\textbf{Limitations of pixel-based gradient estimation.}
Although the aforementioned procedure enables the back-propagation of gradients, it suffers from some major issues. First, the change of the contents of $\tilde{\mathbf{V}}_t$ is achieved by varying the pixel values, each of which is determined only by few adjacent pixels in $\mathbf{V}$. Such a limited source of information may be too local to reflect the semantic-level changes when the location of $\tilde{\mathbf{V}}_t$ varies. However, during training, the model should ideally be informed with how the semantical contents of $\tilde{\mathbf{V}}_t$ will change with the different outputs of $\bm{\pi}$, as we hope to localize the most informative parts of the original videos. Second, performing interpolation along the temporal dimension amounts to mixing the adjacent frames in the pixel space, which is empirically observed to degrade the generalization performance of the model. We tentatively attribute this to the violation of the i.i.d. assumption, since the frames are not mixed at test time.

\begin{figure}[!t]
    \begin{center}
    \centerline{\includegraphics[width=0.925\columnwidth]{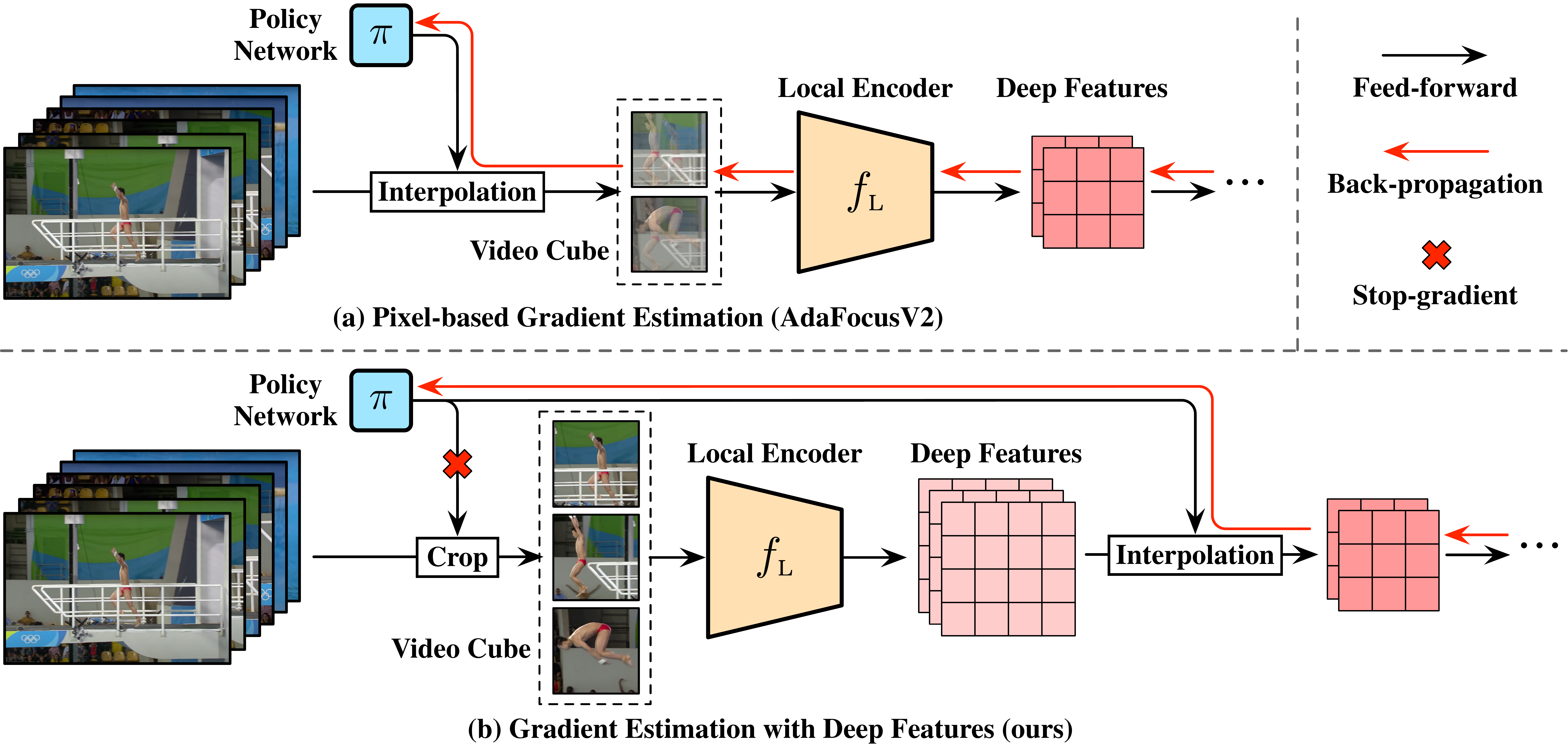}}
    \caption{\textbf{Comparisons of the gradient estimation with image pixels and deep features.} The back-propagation process of the policy network $\bm{\pi}$ is illustrated with \textcolor{red}{\textbf{red arrows}}. AdaFocusV3 proposes to guide the learning of $\bm{\pi}$ with deep features. In comparison with the pixel-based approach proposed in AdaFocusV2, our method not only provides more direct semantic-level supervision signals for $\bm{\pi}$, but also avoids the issue that the adjacent frames are mixed in the pixel space during training.
    \label{fig:grad_estimate}
    }
    \end{center}
\end{figure}

\textbf{Estimating gradients with deep features.}
To alleviate the problems caused by the pixel-based gradient estimation, we propose a novel deep feature based approach to guide the training of the policy network $\bm{\pi}$. Our main motivation here is to enable the outputs of $\bm{\pi}$ to have a direct influence on the feature maps that are extracted by $f_{\textnormal{L}}$. These deep representations are known to excel at capturing the task-relevant semantics of the inputs \cite{bengio2013better,upchurch2017deep,wang2019implicit}. Once the gradients can explicitly tell $\bm{\pi}$ how the deep features corresponding to $\tilde{\mathbf{V}}_t$ will be affected by its outputs, $\bm{\pi}$ will receive direct supervision signals on whether $\tilde{\mathbf{V}}_t$ contains valuable contents in terms of the recognition task.

To implement this idea, assume that ${\bm{e}}_{t}\!\in\!\R^{H^{\textnormal{f}}\times W^{\textnormal{f}}\times T^{\textnormal{f}}}$ denotes the feature maps of $\tilde{\mathbf{V}}_t$ produced by $f_{\textnormal{L}}$. During training, we crop a slightly larger cube $\tilde{\mathbf{V}}_t'$ at the location of $\tilde{\mathbf{V}}_t$ from the original video $\mathbf{V}$, such that the size of the feature maps ${\bm{e}}_{t}'$ corresponding to $\tilde{\mathbf{V}}_t'$ will exactly be $(H^{\textnormal{f}}\!+\!1)\!\times\!(W^{\textnormal{f}}\!+\!1)\!\times\!(T^{\textnormal{f}}\!+\!1)$. Note that here we deactivate the gradient computation and hence the back-propagation through Eqs. (\ref{eq:trilinear}-\ref{eq:bp_2}) will not happen. Then we obtain ${\bm{e}}_{t}$ by performing trilinear interpolation at the centre of ${\bm{e}}_{t}'$. Formally, the value of any element ${e}^t_{i,j,k}$ in ${\bm{e}}_{t}$ can be solved by
\begin{equation}
    \label{eq:trilinear_feature}
    {e}^t_{i,j,k} = \textnormal{Trilinear}\left({\bm{e}}_{t}', ({x}({e}^t_{i,j,k}), {y}({e}^t_{i,j,k}), {z}({e}^t_{i,j,k}))\right).
\end{equation}
Herein, ${x}({e}^t_{i,j,k})$, ${y}({e}^t_{i,j,k})$ and ${z}({e}^t_{i,j,k})$ are the coordinates in ${\bm{e}}_{t}'$ corresponding to ${e}^t_{i,j,k}$. They are given by:
\begin{equation}
    \label{eq:stop_grad}
    ({x}({e}^t_{i,j,k}), {y}({e}^t_{i,j,k}), {z}({e}^t_{i,j,k})) = (\tilde{x}^t_{\textnormal{c}}, \tilde{y}^t_{\textnormal{c}}, \tilde{z}^t_{\textnormal{c}}) - \textnormal{StopGradient}(\tilde{x}^t_{\textnormal{c}}, \tilde{y}^t_{\textnormal{c}}, \tilde{z}^t_{\textnormal{c}}) + \bm{o}_{i,j,k},
\end{equation}
where $\bm{o}_{i,j,k}$ is a vector from $(0, 0, 0)$ to the location of ${e}^t_{i,j,k}$, and it is only conditioned on $i, j, k$ and the shape of ${e}^t_{i,j,k}$. When calculating the gradients with Eq. (\ref{eq:stop_grad}), the infinitesimal change on the outputs of $\bm{\pi}$ (\emph{i.e.}, $\tilde{x}^t_{\textnormal{c}}, \tilde{y}^t_{\textnormal{c}}, \tilde{z}^t_{\textnormal{c}}$) will directly act on the feature maps ${\bm{e}}_{t}$. To sum up, in the feed-forward process, we obtain the enlarged cube $\tilde{\mathbf{V}}_t'$ in the pixel space based on outputs of $\bm{\pi}$, while for back-propagation, the supervision signals for $\bm{\pi}$ come from the semantic-level deep feature space. In addition, it is noteworthy that only the training process is modified. At test time, we crop $\tilde{\mathbf{V}}_t$ and activate $f_{\textnormal{L}}$ as stated in Section \ref{sec:overview}.

\subsection{Implementation Details}
\label{sec:implementation_details}

\textbf{Adaptive inference.}
As mentioned in Section \ref{sec:overview}, the inference procedure can be terminated once the model is already able to produce a sufficiently reliable prediction. Consequently, the computation is unevenly allocated across ``easy'' and ``hard'' videos, improving the overall efficiency. In implementation, after processing the inputs $\{\tilde{\mathbf{V}}_1, \dots, \tilde{\mathbf{V}}_2\}$, the classifier $f_{\textnormal{C}}$ integrates all the acquired information and outputs a softmax prediction $\bm{p}_t$. We compare the entropy of $\bm{p}_t$ (\emph{i.e.}, $-\sum\nolimits_{j}p_{ij}\log p_{ij}$) with a threshold $\eta_t$. The inference will be terminated with $\bm{p}_t$ if we have $-\sum\nolimits_{j}p_{ij}\log p_{ij} \leq \eta_t$. The thresholds $\{\eta_1, \eta_2, \ldots\}$ can be solved on the validation set $\mathcal{D}_{\textnormal{val}}$, under the principle of maximizing the accuracy with a limited computational budget $B>0$:
\begin{equation}
    \label{eq:thres}
    \mathop{\textnormal{maximize}}_{\eta_1, \eta_2, \ldots}\ \textnormal{Accuracy}(\eta_1, \eta_2, \ldots|\mathcal{D}_{\textnormal{val}})\ \ \ \ 
    \textnormal{s.t.}\ \textnormal{FLOPs}(\eta_1, \eta_2, \ldots|\mathcal{D}_{\textnormal{val}})\leq B.
\end{equation}
As a matter of fact, by varying the value of $B$, a group of $\{\eta_1, \eta_2, \ldots\}$ can be obtained, corresponding to a variety of computational constraints. The computational cost of our proposed AdaFocusV3 framework can be online adjusted without additional training by simply adapting these termination criterion. In this paper, we solve problem (\ref{eq:thres}) following the methodology proposed in \cite{huang2017multi,NeurIPS2020_7866,wang2021not}.

\textbf{Glance step.}
As aforementioned, AdaFocusV3 first takes a glance at the input video with the light-weighted global encoder $f_{\textnormal{G}}$. Although being cheap and coarse, the global representations obtained from $f_{\textnormal{G}}$ may have learned certain discriminative features of the inputs. To this end, in addition to feeding them into the policy network $\bm{\pi}$, we also activate classifier $f_{\textnormal{C}}$ to produce a quick prediction $\bm{p}_0$ only using this global information, aiming to facilitate efficient feature reuse.










\section{Experiment}
\label{sec:experiment}

In this section, we present a comprehensive experimental comparison between our proposed AdaFocusV3 and state-of-the-art efficient video recognition frameworks. AdaFocusV3 significantly outperforms these competitive baselines in terms of computational efficiency. We also demonstrate that AdaFocusV3 can be deployed on the basis of recently processed light-weighted deep networks, and further improve their efficiency. The analytical results including the ablation study and visualization are provided to give additional insights into our method. 



\textbf{Datasets.}
Our experiments are based on six large-scale video recognition benchmark datasets, \emph{i.e.}, ActivityNet \cite{caba2015activitynet}, FCVID \cite{TPAMI-fcvid}, Mini-Kinetics \cite{kay2017kinetics,wu2019liteeval}, Something-Something (Sth-Sth) V1\&V2 \cite{goyal2017something} and Diving48 \cite{li2018resound}. The official training-validation split is adopted for all of them. Note that these datasets are widely used in the experiments of a considerable number of recently proposed baselines. We select them for a reasonable comparison with current state-of-the-art results. Due to the spatial limitations, the detailed introductions on datasets and data pre-processing are deferred to Appendix A.

\textbf{Metrics.}
The offline video recognition is considered, where a single prediction is required for each video. Following the common practice on the aforementioned datasets, we evaluate the performance of different methods on ActivityNet and FCVID via mean average precision (mAP), while the top-1 accuracy is adopted on Mini-Kinetics, Sth-Sth V1\&V2 and Diving48. In addition, we measure the theoretical computational cost of the models using the average number of multiply-add operations required for processing a video (\emph{i.e.}, GFLOPs/Video). To compare the practical inference speed of different methods, we report the maximal throughput (Videos/s) when a given hardware (\emph{e.g.}, GPU) is saturated.


\subsection{Comparisons with State-of-the-art Baselines}

\textbf{Baselines.}
We first compare AdaFocusV3 with a variety of recently proposed approaches that focus on improving the efficiency of video recognition. The results on ActivityNet, FCVID and Mini-Kinetics are provided. In addition to the previous versions of AdaFocus, the following baselines are included, \emph{i.e.}, LiteEval \cite{wu2019liteeval}, SCSampler \cite{korbar2019scsampler}, ListenToLook \cite{gao2020listen}, AR-Net \cite{meng2020ar}, AdaFrame \cite{wu2019adaframe}, VideoIQ \cite{sun2021dynamic}, and OCSampler \cite{lin2022ocsampler}. An introduction of them is presented in Appendix B.

\textbf{Implementation details.}
For a fair comparison, the design of backbone networks follows the common setups of the baselines. A MobileNet-V2 \cite{sandler2018mobilenetv2} and a ResNet-50 \cite{he2016deep} are used as the global encoder $f_{\textnormal{G}}$ and local encoder $f_{\textnormal{L}}$ in AdaFocusV3. The architecture of the classifier $f_{\textnormal{C}}$ follows from \cite{ghodrati2021frameexit,wang2021adafocus}. In addition, given that the videos in ActivityNet, FCVID and Mini-Kinetics are relatively long, previous works have revealed that a single uniformly sampled frame is sufficient to represent the information within the adjacent video clip \cite{wu2019adaframe,ghodrati2021frameexit,lin2022ocsampler}. Hence, we set the cube size of AdaFocusV3 to be $96\!\times\!96\times\!1$ and $128\!\times\!128\!\times\!1$, which enables our method to flexibly perform spatial-temporal dynamic computation in a frame-wise fashion. The maximum number of video cubes is set to 18. During inference, we remove the frames which have been included by previous cubes. The training configurations can be found in Appendix C.

\begin{figure}[!t]
    \begin{center}
    \centerline{\includegraphics[width=\columnwidth]{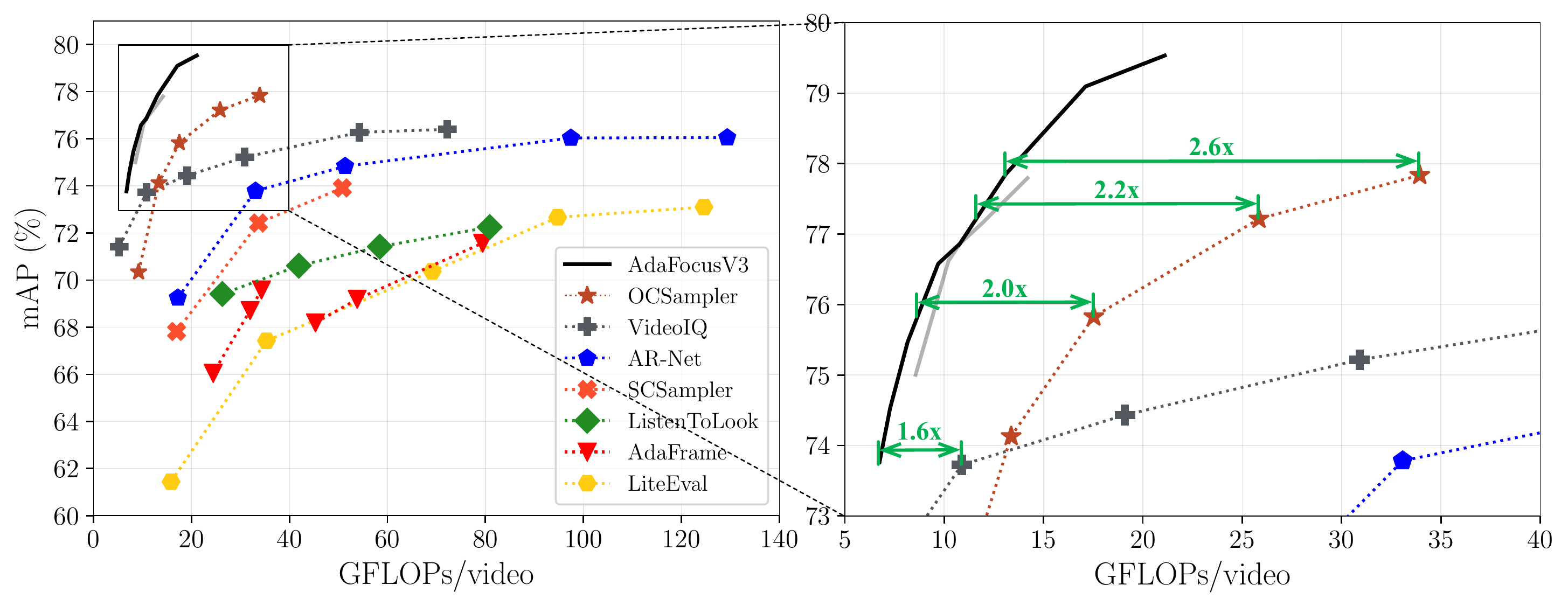}}
    \caption{\textbf{AdaFocusV3 v.s. state-of-the-art baselines on ActivityNet in terms of computational efficiency.} Note that the computational cost of AdaFocusV3 can be adjusted online (\emph{i.e.}, switching within each black curve without additional training).
    \label{fig:vs_sota}
    }
    \end{center}
\end{figure}

\textbf{Main results.}
We report the performance of AdaFocusV3 and the baselines in Table \ref{tab:main_result_table}. The results of our method are presented when its accuracy or computational cost reaches the best records of all the baselines. It is clear that AdaFocusV3 outperforms other efficient video recognition frameworks dramatically. For example, it improves the mAP by $\sim2.6\!-\!3.2\%$ compared to the competitive OCSampler \cite{lin2022ocsampler} algorithm on ActivityNet and FCVID with the same GFLOPs. A more comprehensive comparison is depicted in Figure \ref{fig:vs_sota}, where we vary the computational budget and plot the ActivityNet mAP v.s. GFLOPs relationships of both AdaFocusV3 and the variants of baselines. The two black curves correspond to adopting the two sizes of video cubes. One can observed that AdaFocusV3 yields a considerably better accuracy-efficiency trade-off consistently. It reduces the demands of computation by $\sim1.6\!-\!2.6\times$ on top of the strongest baselines without sacrificing mAP.


\begin{table*}[t]
  \centering
  \begin{footnotesize}
  \caption{\textbf{Performance of AdaFocusV3 and the baselines on three benchmark datasets}. The comparisons are based on the same backbones, where MN2/RN denotes MobileNet-V2/ResNet and $\dagger$ represents adding TSM \cite{lin2019tsm}. GFLOPs refer to the average computational cost for processing each video. The best two results are \textbf{bold-faced} and \underline{underlined}, respectively. \textcolor{blue}{Blue numbers} are obtained on top of the strongest baselines. 
  }
  \label{tab:main_result_table}
  \setlength{\tabcolsep}{0.5mm}{
  \renewcommand\arraystretch{1}
  \resizebox{\columnwidth}{!}{
  \begin{tabular}{c|c|c|cccccc}
  \toprule
  \multirow{2}{*}{Methods} & \multirow{2}{*}{Published on}& \multirow{2}{*}{Backbones}  & \multicolumn{2}{c}{ActivityNet} &  \multicolumn{2}{c}{FCVID} & \multicolumn{2}{c}{Mini-Kinetics}  \\
  &&& mAP & GFLOPs & mAP &  GFLOPs & Top-1 Acc. &  GFLOPs \\
  
  \midrule
  LiteEval \cite{wu2019liteeval} & {NeurIPS'19} &MN2+RN  & 72.7\% & 95.1 & 80.0\% & 94.3  & 61.0\% & 99.0 \\
  SCSampler \cite{korbar2019scsampler} & {ICCV'19} &MN2+RN & 72.9\%  & 42.0 & 81.0\% & 42.0  & 70.8\%  & 42.0 \\
  ListenToLook \cite{gao2020listen} & {CVPR'20} &MN2+RN & 72.3\%  & 81.4 & -- & --  & -- & --  \\
  AR-Net \cite{meng2020ar} & {ECCV'20} &MN2+RN & 73.8\% & 33.5 & 81.3\% & 35.1  & 71.7\% & 32.0  \\
  AdaFrame \cite{wu2019adaframe} & {T-PAMI'21} &MN2+RN & 71.5\% & 79.0 & 80.2\% & 75.1  & -- & --  \\
  VideoIQ \cite{sun2021dynamic} & {ICCV'21} &MN2+RN & 74.8\% &  28.1 & {82.7\%} & {27.0}  & 72.3\% &  20.4 \\
  OCSampler \cite{lin2022ocsampler} & {CVPR'22} &MN2$^\dagger$+RN & \underline{76.9\%} &  \underline{21.7} & \underline{82.7\%} & \underline{26.8}  & \underline{72.9\%} &  \underline{17.5} \\
  \midrule
  \multirow{2}{*}{AdaFocusV3}  & -- &MN2+RN  & \underline{76.9\%} & \textbf{10.9}\textcolor{blue}{$_{\downarrow 2.0\!\times}$} & \underline{82.7\%} & \textbf{7.8}\textcolor{blue}{$_{\downarrow 3.4\!\times}$} & \underline{72.9\%} & \textbf{8.6}\textcolor{blue}{$_{\downarrow 2.0\!\times}$} \\
    & -- &MN2+RN  & \ \textbf{79.5\%}\textcolor{blue}{$_{\uparrow 2.6\%}$} & \underline{21.7} & \textbf{85.9\%}\textcolor{blue}{$_{\uparrow 3.2\%}$} & \underline{26.8} & \textbf{75.0\%}\textcolor{blue}{$_{\uparrow 2.1\%}$} & \underline{17.5} \\
  \bottomrule
  \end{tabular}}}
  \end{footnotesize}
\end{table*}

\begin{table*}[!t]
    \centering
    \begin{footnotesize}
    \caption{\textbf{Comparisons of AdaFocusV3 and AdaFocusV2+.} The latter differs from AdaFocusV3 only in that it models the spatial and temporal redundancy independently in isolated modules. We report the computational cost of the two methods when they reach the same accuracy.}
    \label{tab:v3_vs_v2}
    \setlength{\tabcolsep}{4mm}{
    \renewcommand\arraystretch{1}
    \resizebox{0.65\columnwidth}{!}{
    \begin{tabular}{cccc}
    \toprule
    \multirow{2}{*}{Dataset} & \multirow{2}{*}{mAP}& \multicolumn{2}{c}{Computational Cost (GFLOPs/video)} \\
    && AdaFocusV2+ &  AdaFocusV3 (ours) \\
    \midrule
    \multirow{3}{*}{ActivityNet} & 77.0\% & 14.0 & \textbf{11.1} (\textcolor{blue}{$\downarrow$1.3$\times$}) \\
    & 78.0\% & 17.6 & \textbf{13.6} (\textcolor{blue}{$\downarrow$1.3$\times$}) \\
    & 79.0\% & 24.4 & \textbf{16.8} (\textcolor{blue}{$\downarrow$1.5$\times$}) \\
    \midrule
    \multirow{3}{*}{FCVID} & 83.0\% & 10.8 & \textbf{8.3} (\textcolor{blue}{$\downarrow$1.3$\times$}) \\
    & 84.0\% & 14.2 & \textbf{10.5} (\textcolor{blue}{$\downarrow$1.4$\times$}) \\
    & 85.0\% & 24.4 & \textbf{14.4} (\textcolor{blue}{$\downarrow$1.7$\times$}) \\
    \bottomrule
    \end{tabular}}}
    \end{footnotesize}
\end{table*}

\begin{table*}[!t]
  \centering
  \begin{footnotesize}
  \caption{\textbf{Performance of AdaFocusV3 on top of TSM}. MN2, R18/R34/R50 and BN-Inc. refer to MobileNet-V2, ResNet-18/34/50 and BN-Inception, respectively. TSM+ denotes the augmented TSM baseline with the same backbone network architecture as our method. The best results are \textbf{bold-faced}.}
  \label{tab:sthsth}
  \setlength{\tabcolsep}{0.5mm}{
  \renewcommand\arraystretch{1}
  \resizebox{\columnwidth}{!}{
  \begin{tabular}{ccccccccc} 
  \toprule
  \multirow{2}{*}{{Method}} & \multirow{2}{*}{{Backbones}}    & \multicolumn{2}{c}{{Sth-Sth V1}}  & \multicolumn{2}{c}{{Sth-Sth V2}}& Throughput (Sth-Sth V1)\\ 
  && \footnotesize{{\ \ Top-1 Acc.}} & \footnotesize{{GFLOPs}} &  \footnotesize{{Top-1 Acc.}} & \footnotesize{{GFLOPs}}  & \scriptsize{(NVIDIA 3090 GPU, bs=128)}\\  
  \midrule
  TSN \cite{wang2016temporal} & R50  & 19.7\% & 33.2 &27.8\% &  33.2   & - \\ 
  AR-Net \cite{meng2020ar} & MN2+R18/34/50  & 18.9\% & 41.4 & -  &  -   & - \\ 
  TRN\textsubscript{RGB/Flow} \cite{zhou2018temporal} & BN-Inc.   & 42.0\% & 32.0 & 55.5\% & 32.0   & - \\
  ECO \cite{zolfaghari2018eco} & BN-Inc.+3DR18  & 39.6\% & 32.0 & - & - & -\\
  STM \cite{jiang2019stm} & R50  & 47.5\% & 33.3 &  - & -   & - \\
  TSM \cite{lin2019tsm}  & R50  & 46.1\% & 32.7 &  59.1\% & 32.7  &  {162.7 Videos/s} \\
  AdaFuse-TSM \cite{meng2021adafuse} & R50  & 46.8\% & 31.5 & 59.8\% & 31.3   & - \\
  \midrule
  TSM+ \cite{lin2019tsm}  & MN2+R50   & 47.0\% & 35.1 & 59.6\% & 35.1 &   { 123.0 Videos/s}   \\
  AdaFocusV2  & MN2+R50  & 47.0\% & {18.5} (\textcolor{blue}{$\downarrow$1.9$\times$}) & 59.6\%  & {18.5} (\textcolor{blue}{$\downarrow$1.9$\times$}) &{ 197.0 Videos/s} (\textcolor{blue}{$\uparrow$1.6$\times$})   \\
  AdaFocusV3  & MN2+R50 & 47.0\%& \textbf{14.0} (\textcolor{blue}{$\downarrow$\textbf{2.5}$\times$}) & 59.6\% & \textbf{15.4} (\textcolor{blue}{$\downarrow$\textbf{2.3}$\times$}) & \ \ \textbf{234.0 Videos/s} (\textcolor{blue}{$\uparrow$\textbf{1.9}$\times$}) \\
  \bottomrule
  \end{tabular}}}
  \end{footnotesize}
\end{table*}

\begin{table}[!t]
\centering
\begin{footnotesize}
\caption{\label{tab:qubesize}\textbf{AdaFocusV3 on top of TSM with varying sizes of video cubes}. MN2/R50 denotes MobileNetV2/ResNet50.  
}
\setlength{\tabcolsep}{1.5mm}{
\renewcommand\arraystretch{1}
\resizebox{0.85\columnwidth}{!}{
\begin{tabular}{c|cl|cc|cc|cc} 
\toprule
\multirow{2}{*}{{Method}} & \multirow{2}{*}{Backbones}  & \multicolumn{1}{c|}{Size of 3D} & \multicolumn{2}{c|}{{Sth-Sth V1}}  & \multicolumn{2}{|c|}{{Sth-Sth V2}}& \multicolumn{2}{|c}{Diving48}\\ 
&&\multicolumn{1}{c|}{Video Cubes}& \footnotesize{{\ \ Acc.}} & \footnotesize{{GFLOPs}} &  \footnotesize{{\ Acc.}} & \footnotesize{{GFLOPs}}   &  \footnotesize{{\ Acc.}} & \footnotesize{{GFLOPs}}\\  
\midrule
TSM \cite{lin2019tsm}  & { {R50}} & \multicolumn{1}{c|}{--}  & \ 46.1\% & 32.7 &  \ 59.1\% & 32.7  &  \ 77.4\% & 32.7 \\
\midrule
TSM+ \cite{lin2019tsm}   & { {MN2+R50}} & \ \ $224^2\!\times\!16$   & \ 47.0\% & 35.1 & \ 59.6\% & 35.1 &   \ 80.4\%  & 35.1  \\
AdaFocusV3  & { {MN2+R50}}  & \ \ $128^2\!\times\!8$ & \ 46.1\% & 14.0  & \ 58.5\% & 15.4  & \ 79.3\% & 6.2 \\
\multirow{1}{*}{AdaFocusV3}  & { {MN2+R50}}  & \multirow{1}{*}{\ \ $128^2\!\times\!12$} 
& \multirow{1}{*}{\ \ \!\textbf{47.0\%}} & \multirow{1}{*}{{\textbf{14.0}}}  
& \multirow{1}{*}{\ \ \!\textbf{59.6\%}} & \multirow{1}{*}{{\textbf{15.4}}}
& \multirow{1}{*}{\ \ \!\textbf{80.4\%}} & \multirow{1}{*}{{\textbf{6.2}}} \\
AdaFocusV3  & { {MN2+R50}}  & \ \ $128^2\!\times\!16$ & \ 47.0\% & 15.9  & \ 59.6\% & 17.2  & \ 80.4\% & 7.7 \\
\bottomrule
\end{tabular}}}
\end{footnotesize}
\end{table}

\textbf{Comparisons of the unified and separate modeling of spatial and temporal redundancy.}
We further empirically validate whether a unified framework for spatial-temporal dynamic computation indeed outperforms the isolatedly-modeling counterpart. This is achieved by comparing AdaFocusV3 with AdaFocusV2+ \cite{wang2021adafocus}. The latter has the same network architecture and training procedure as AdaFocusV3, while the only difference lies in that AdaFocusV2+ models spatial and temporal redundancy independently in isolated modules. The results in Table \ref{tab:v3_vs_v2} illustrate that our proposed unified framework efficiently reduce the computational cost (by $\sim1.3\times$) with preserved accuracy.

\subsection{Deploying on Top of Light-weighted Models}

\textbf{Setups.}
In this subsection, we implement our framework on top of a representative efficient video analysis network, ConvNets with temporal shift module (TSM) \cite{lin2019tsm}, to demonstrate its effectiveness on recently proposed light-weighted models. The network architecture is not changed except for adding TSM to $f_{\textnormal{G}}$ and $f_{\textnormal{L}}$. We uniformly sample 8 video frames for $f_{\textnormal{G}}$. Unless otherwise specified, the size of the video cubes for $f_{\textnormal{L}}$ is set to be $128\!\times\!128\!\times\!12$, while the maximum number of cubes is 2. The training configurations can be found in Appendix C.


\begin{table}[t]
  \centering
  \begin{footnotesize}
  \caption{\textbf{Ablation study on the learned policy and training algorithm.}}
  \label{tab:abl_study}
  \setlength{\tabcolsep}{2mm}{
  \renewcommand\arraystretch{1}
  \resizebox{0.825\columnwidth}{!}{
  \begin{tabular}{ccc|cccc}
  \toprule
  \multirow{3}{*}{\shortstack{Gradient\\ Estimation \\ with Deep Features}} &\multirow{3}{*}{\shortstack{Learned \\ Temporal-dynamic\\ Policy}}&\multirow{3}{*}{\shortstack{Learned \\ Spatial-dynamic\\ Policy}} & \multicolumn{4}{|c}{ActivityNet mAP after processing} \\
  &&& \multicolumn{4}{|c}{ $t$ video cubes (\emph{i.e.}, with $\bm{p}_t$)} \\
  &&& $t$=1 & $t$=2 & $t$=4 & $t$=8 \\
  \midrule
  --& \textcolor{black}{\xmark} (random) & \textcolor{black}{\xmark} (random) &\ 44.6\% & 57.6\% & 67.4\% & 73.7\%  \\
  \textcolor{black}{\cmark}& \textcolor{black}{\cmark} & \textcolor{black}{\xmark} (random) &\ 50.1\% & 61.3\% & 69.4\% &  74.6\%  \\
  \textcolor{black}{\cmark}& \textcolor{black}{\xmark} (random) & \textcolor{black}{\cmark} &\ 45.5\% & 58.2\% & 68.5\% & 74.6\%  \\
  \textcolor{black}{\xmark} (pixel-based)& \textcolor{black}{\cmark} & \textcolor{black}{\cmark} &\ 50.6\% & 61.0\% & 69.5\% & 75.0\%  \\
  \midrule
  \textcolor{black}{\cmark}& \textcolor{black}{\cmark} & \textcolor{black}{\cmark} & \ \textbf{51.9\%} & \textbf{62.3\%} & \textbf{70.6\%} & \textbf{75.4\%}  \\
  \bottomrule
  \end{tabular}}}
  \end{footnotesize}
\end{table}

\begin{figure}[!t]
  \begin{center}
  \centerline{\includegraphics[width=0.8\columnwidth]{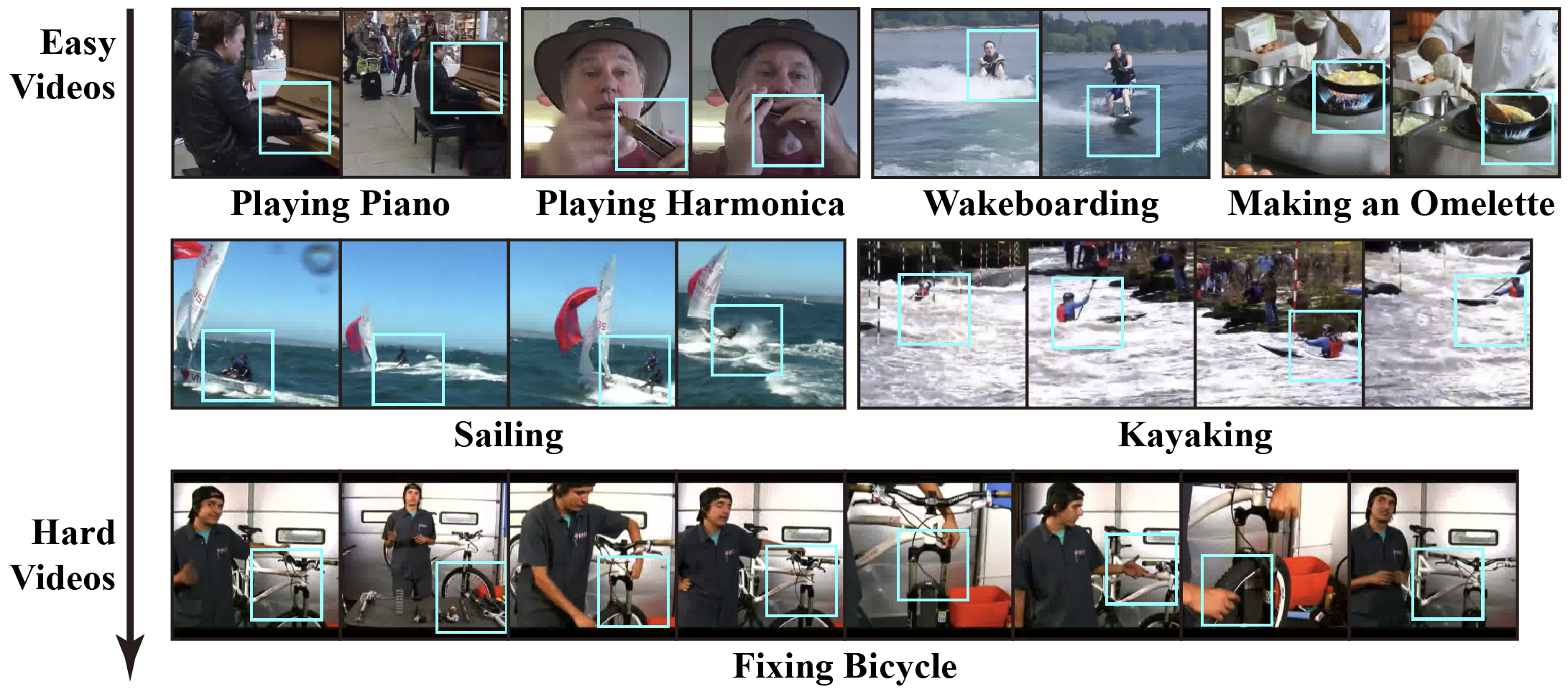}}
  \caption{\textbf{Visualization results on ActivityNet.} Easy and hard videos refer to the samples that need a small or large number of video cubes for being correctly recognized.
  \label{fig:visualization}
  }
  \end{center}
\end{figure}

\begin{figure}[!t]
  \begin{center}
  \centerline{\includegraphics[width=0.8\columnwidth]{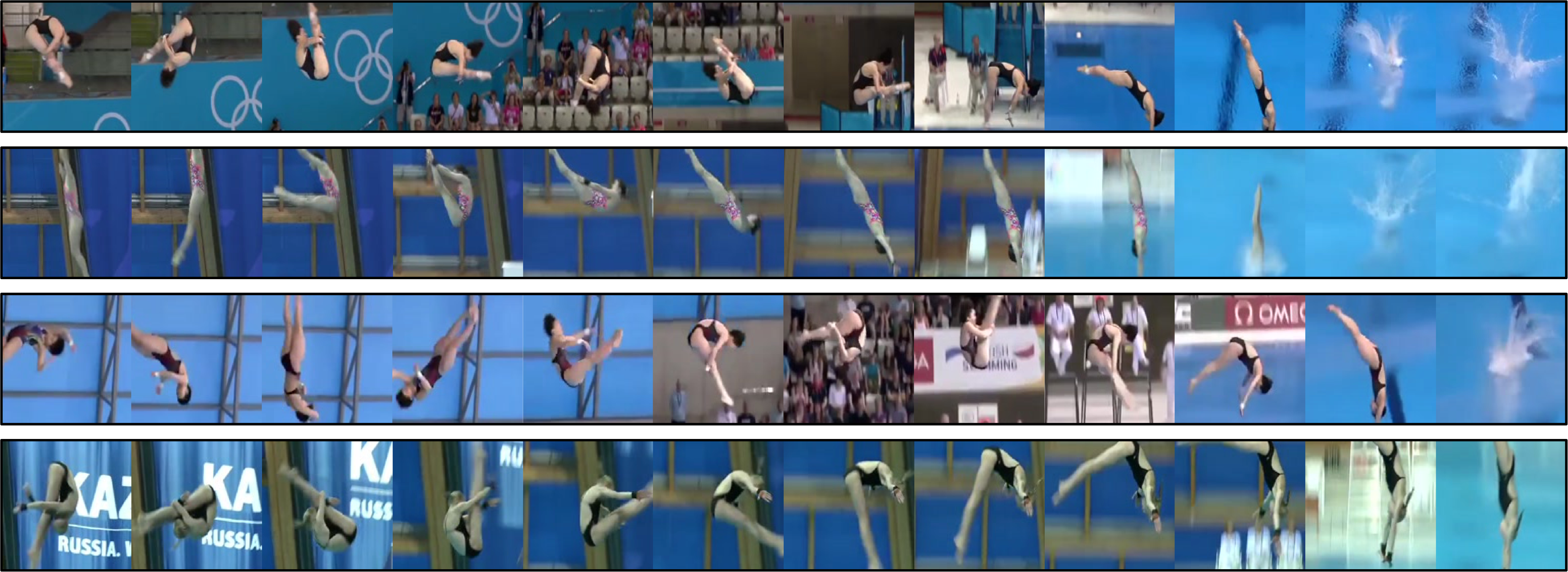}}
  \caption{\textbf{Visualization results on Diving48.} We present the examples of the task-relevant 3D video cubes localized by AdaFocusV3.
  \label{fig:vis_diving}
  }
  \end{center}
\end{figure}

\textbf{Results on Sth-Sth V1\&V2}
are presented in Table \ref{tab:sthsth}. For fair comparisons, following \cite{wang2021adafocus}, we augment the vanilla TSM by adopting the same two backbone networks as ours, which is named as TSM+. It can be observed that AdaFocusV3 saves the computational cost by up to $2.5\times$ compared with TSM+ when reaching the same Top-1 Acc. We also test the practical efficiency of our method on GPU devices, where the videos are fed into the model in batches, and the samples that meet the early-termination criterion will be output at each exit. The results indicate that the real speedup of AdaFocusV3 is significant as well.

\textbf{Results with varying cube sizes}
are summarized in Table \ref{tab:qubesize}. We find that reducing the cube size in the temporal dimension ($128^2\!\times\!12\!\to\!128^2\!\times\!8$) significantly degrades the accuracy on Sth-Sth/Diving48. Also interestingly, larger cubes like $128^2\!\times\!16$ may weaken the temporal adaptiveness, yielding inferior computational efficiency. Another interesting observation is that on Diving48, AdaFocusV3 dramatically reduces the computational cost (by $5.7\!\ \!\times$) on top of the baseline (TSM+) with the same accuracy. This phenomenon may be explained by that AdaFocusV3 will be more effective in less biased scenarios, where the background is less informative, and thus it is more sensible to attend to the task-relevant foreground parts of videos.

\subsection{Analytical Results}

\textbf{Ablation study.}
In Table \ref{tab:abl_study}, we test ablating the cube selection policy and the gradient estimation technique in AdaFocusV3.  The results on ActivityNet with the cube size of $128^2\!\times\!1$ are presented. For a clean comparison, here we deactivate the glance step as well as the early-termination algorithm, and report the mAP corresponding to processing a fixed number of video cubes. One can observe that our learned policy considerably outperforms the random baselines in terms of either spatial or temporal dimensions. In addition, our gradient estimation technique based on deep features significantly outperforms the pixel-based counterpart proposed by AdaFocusV2 \cite{wang2021adafocus}.


\textbf{Visualization.} 
In Figure \ref{fig:visualization}, the visualization on ActivityNet with the cube size of $96^2\!\times\!1$ is presented. The blue boxes indicate the video cubes localized by AdaFocusV3. It is shown that our method adaptively attends to the task-relevant parts of the video, such as the person, the piano and the bicycle. Figure \ref{fig:vis_diving} shows the examples of video cubes on Diving48. AdaFocusV3 can adaptively capture the actions of the diving athletes -- the crucial contents for recognition.

\section{Conclusion}

In this paper, we proposed an AdaFocusV3 framework which enables the unified formulation of the efficient spatial-temporal dynamic computation in video recognition. AdaFocusV3 is trained to adaptively identify and attend to the most task-relevant video cubes in the 3D space formed by frame height/width and the time. The number of these cubes are determined conditioned on each sample under the dynamic early-termination paradigm. Extensive empirical results on six benchmark datasets validate the state-of-the-art performance of our model in terms of computational efficiency. Our work may open new avenues for the simultaneous modeling of spatial and temporal redundancy in video recognition.

\section*{Acknowledgements}

This work is supported in part by National Key R\&D Program of China\\ (2020AAA0105200), the National Natural Science Foundation of China under Grants 62022048, Guoqiang Institute of Tsinghua University and Beijing Academy of Artificial Intelligence. We also appreciate the generous donation of computing resources by High-Flyer AI.


{\small
\bibliographystyle{splncs04}
\bibliography{egbib}
}

\appendix
\newpage

\section*{Appendix}


\section{Experimental Setups}

\textbf{Datasets.}
Our experiments are based on six large-scale video recognition benchmark datasets, \emph{i.e.}, ActivityNet \cite{caba2015activitynet}, FCVID \cite{TPAMI-fcvid}, Mini-Kinetics \cite{kay2017kinetics,wu2019liteeval}, Something-Something (Sth-Sth) V1\&V2 \cite{goyal2017something} and Diving48 \cite{li2018resound}. The official training-validation split is adopted for all of them. Note that these datasets are widely used in the experiments of a considerable number of recently proposed baselines. We select them for a reasonable comparison with current state-of-the-art results.
\begin{itemize}
    \item ActivityNet \cite{caba2015activitynet} contains the videos of 200 human action categories. It includes 10,024 training videos and 4,926 validation videos. The average duration is 117 seconds.    
    \item FCVID \cite{TPAMI-fcvid} includes 45,611 training videos and validation 45,612 videos. The data is annotated into 239 classes. The average duration is 167 seconds.    
    \item Mini-Kinetics is a subset of the Kinetics \cite{kay2017kinetics} dataset. It contains include 200 randomly selected classes of videos, with 121k videos for training and 10k videos for validation. The average duration is around 10 seconds \cite{kay2017kinetics}. We establish it following \cite{wu2019liteeval,meng2020ar,sun2021dynamic,lin2022ocsampler}.
    \item Something-Something (Sth-Sth) V1\&V2 \cite{goyal2017something} datasets contain 98k and 194k videos respectively. Both of them are labeled with 174 human action categories. The average duration is 4.03 seconds.
    \item  Diving48 \cite{li2018resound} is a fine-grained video dataset of competitive diving, consisting of $\sim$18k trimmed video clips of 48 unambiguous dive sequences.
\end{itemize}

\textbf{Data pre-processing.}
We uniformly sample 18 frames from each video on ActivityNet, FCVID and Mini-Kinetics, while sampling 8/16 frames on Sth-Sth and Diving48. These configurations are determined on the validation set for a favorable accuracy-efficiency trade-off. The data augmentation pipeline in \cite{lin2019tsm,meng2020ar,sun2021dynamic,Wang_2021_ICCV,wang2021adafocus} is adopted. Specifically, the frames of training data is randomly scaled and cropped into 224$\times$224 images. On all the datasets except for Sth-Sth V1\&V2 and Diving48, the random flipping is performed as well. At test time, all the frames are resized to 256$\times$256 and centre-cropped to 224$\times$224.

\section{Baselines}

\textbf{Baselines.}
We compare AdaFocusV3 with a variety of recently proposed approaches that focus on improving the efficiency of video recognition. The results on ActivityNet, FCVID and Mini-Kinetics are provided. In addition to the previous versions of AdaFocus, the following baselines are included.
\begin{itemize}
  \item LiteEval \cite{wu2019liteeval} dynamically activates coarse and fine LSTM networks conditioned on the importance of each frame.
  \item SCSampler \cite{korbar2019scsampler} is an efficient framework to select salient video clips or frames. The implementation in \cite{meng2020ar} is adopted.
  \item ListenToLook \cite{gao2020listen} searches for the task-relevant video frames by leveraging audio information. We adopt the image-based variant introduced in their paper for fair comparisons, since we do not use the audio of videos.
  \item AR-Net \cite{meng2020ar} processes the frames with different resolutions based on their relative importance.
  \item AdaFrame \cite{wu2019adaframe} adaptively identifies the informative frames from the videos with reinforcement learning. 
  \item VideoIQ \cite{sun2021dynamic} learns to process each frame with different precision based the importance in terms of video recognition. 
  \item OCSampler \cite{lin2022ocsampler} is a one-stage framework that learns to represent the video with several informative frames with reinforcement learning.
\end{itemize}

\section{Training Details}

On ActivityNet, FCVID and Mini-Kinetics, the training of AdaFocusV3 exactly follows the same end-to-end training pipeline as AdaFocusV2 \cite{wang2021adafocus}. On Something-Something (Sth-Sth) V1\&V2 and Diving48, we first train the two deep encoders and the classifier with a random policy, and then train the policy network isolatedly. We find that this two-stage training pipeline yields a better performance, while its training cost is approximately the same as its end-to-end training counterpart.

\section{More Results}

\begin{figure}[h]
    \begin{minipage}[t]{\linewidth}
    \centering
    \includegraphics[width=0.5\textwidth]{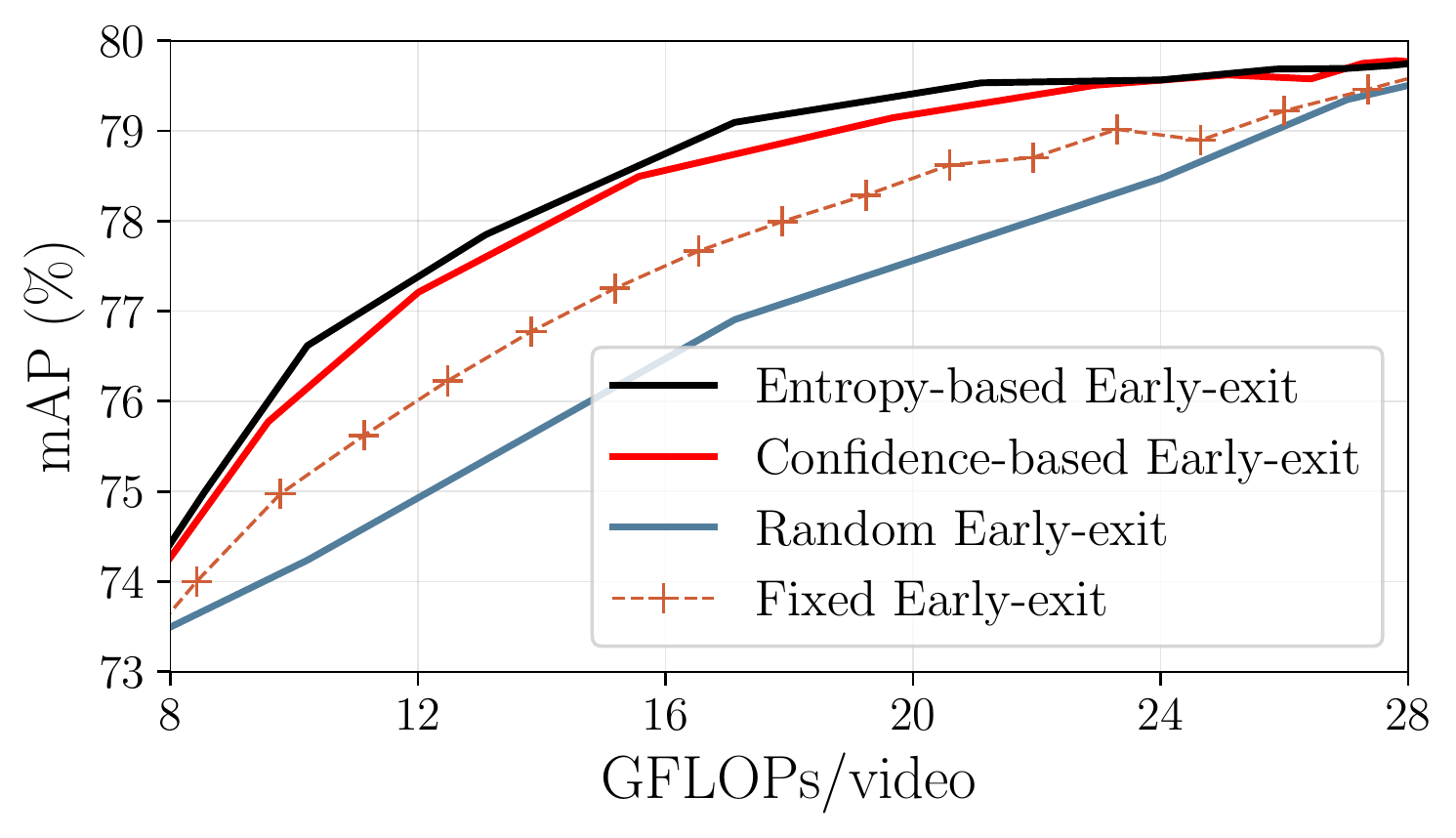}	
    \caption{
      \textbf{Ablation study on early-termination algorithm.} 
    \label{fig:abl_early}}  
    \end{minipage}
\end{figure}

\textbf{Effectiveness of the early-termination algorithm} 
is validated in Figure \ref{fig:abl_early}. The results on ActivityNet with the cube size of $128\!\times\!128\!\times\!1$ are presented. Three variants are considered: (1) adaptive early-exit with prediction confidence; (2) random early-exit with the same exit proportion as AdaFocusV3; (3) early-exit with fixed cube number. Our entropy-based mechanism shows the best performance.

\end{document}